\documentclass[10pt, paper]{IEEEtran}
\usepackage{amsthm}
\usepackage{stfloats}
\usepackage{graphicx}
\usepackage{booktabs}
\usepackage{float}
\usepackage{amssymb,amsfonts}
\usepackage{listings}
\usepackage{url}
\usepackage{algorithm,algorithmicx}
\usepackage{algpseudocode}
\usepackage{enumitem}
\usepackage{mathrsfs}
\usepackage{amsmath,bm}
\usepackage{amsfonts}
\usepackage{amsmath}
\usepackage[switch]{lineno}
\usepackage{colortbl}
\usepackage{scalerel}
\usepackage{tikz}
\usetikzlibrary{svg.path}

\definecolor{orcidlogocol}{HTML}{A6CE39}
\tikzset{
  orcidlogo/.pic={
    \fill[orcidlogocol] svg{M256,128c0,70.7-57.3,128-128,128C57.3,256,0,198.7,0,128C0,57.3,57.3,0,128,0C198.7,0,256,57.3,256,128z};
    \fill[white] svg{M86.3,186.2H70.9V79.1h15.4v48.4V186.2z}
                 svg{M108.9,79.1h41.6c39.6,0,57,28.3,57,53.6c0,27.5-21.5,53.6-56.8,53.6h-41.8V79.1z M124.3,172.4h24.5c34.9,0,42.9-26.5,42.9-39.7c0-21.5-13.7-39.7-43.7-39.7h-23.7V172.4z}
                 svg{M88.7,56.8c0,5.5-4.5,10.1-10.1,10.1c-5.6,0-10.1-4.6-10.1-10.1c0-5.6,4.5-10.1,10.1-10.1C84.2,46.7,88.7,51.3,88.7,56.8z};
  }
}
\newcommand\orcidicon[1]{\href{https://orcid.org/#1}{\mbox{\scalerel*{
\begin{tikzpicture}[yscale=-1,transform shape]
\pic{orcidlogo};
\end{tikzpicture}
}{|}}}}

\usepackage{hyperref}

\ifCLASSINFOpdf
\else
\fi
\begin{document}
%
\title{Fast Symbolic 3D Registration Solution}
%
%
%

\author{Jin~Wu$^ {\orcidicon{0000-0001-5930-4170}\,}$,~\IEEEmembership{Member,~IEEE,}
        Ming~Liu$^{\orcidicon{0000-0002-4500-238X}\,}$,~\IEEEmembership{Senior~Member,~IEEE,}
        Zebo~Zhou$^ {\orcidicon{0000-0002-3687-7133}\,}$
        and~Rui~Li,~\IEEEmembership{Member,~IEEE}
\thanks{Manuscript received May 29, 2018; revised January 3, 2019; accepted September 17, 2019. Date of publication September 17, 2019; date of current version September 17, 2019. This paper was recommended for publication by Associate Editor D. Liu and Editor Y. Sun upon evaluation of the reviewers’ comments. This research was supported by Shenzhen Science, Technology and Innovation Comission (SZSTI) JCYJ20160401100022706, awarded to Prof. Ming Liu., in part by National Natural Science Foundation of China under the grant of No. 41604025, in part by General Research Fund of Research Grants Council Hong Kong,11210017, and also in part by Early Career Scheme Project of Research Grants Council Hong Kong, 21202816. (Corresponding author: Ming Liu).}
\thanks{J. Wu and M. Liu are with Department of Electrical \& Computer Engineering, Hong Kong University of Science and Technology, Hong Kong, China. e-mail: jin\_wu\_uestc@hotmail.com; eelium@ust.hk}
\thanks{Z. Zhou and R. Li are with University of Electronic Science and Technology of China, Chengdu, China. e-mail: klinsmann.zhou@gmail.com; hitlirui@gmail.com.}
\thanks{Color versions of one or more of the figures in this paper are available online at http://ieeexplore.ieee.org.}
\thanks{Digital Object Identifier 10.1109/TASE.2019.}
}

\maketitle


\begin{abstract}
3D registration has always been performed invoking singular value decomposition (SVD) or eigenvalue decomposition (EIG) in real engineering practices. However, these numerical algorithms suffer from uncertainty of convergence in many cases. A novel fast symbolic solution is proposed in this paper by following our recent publication in this journal. The equivalence analysis shows that our previous solver can be converted to deal with the 3D registration problem. Rather, the computation procedure is studied for further simplification of computing without complex-number support. Experimental results show that the proposed solver does not loose accuracy and robustness but improves the execution speed to a large extent by almost \%50 to \%80, on both personal computer and embedded processor.

\indent \textit{Note to Practitioners}---3D registration usually has large computational burden in engineering tasks. The proposed symbolic solution can directly solve the eigenvalue and its associated eigenvector. A lot of computation resources can then be saved for better overall system performance. The deterministic behavior of the proposed solver also ensures long-endurance stability and can help engineer better design thread timing logic.
\end{abstract}


%
\IEEEpeerreviewmaketitle

\section{Introduction}
%
%
%
%

\IEEEPARstart{M}{otion} estimation from point correspondences is an important technique in robotics \cite{Aghili2016, Wang2015Real, Ye2015}. The point measurements can usually be acquired from laser scanner and camera for accurate relative attitude/position determination \cite{Yang2016, Ye2004}. The methodology behind is called the 3D registration which figures out the rigid transformation consisting of rotation and translation \cite{faber1988orientation}. Yet, this technology is employed for 3D reconstruction of objects by means of multi-directional point-cloud snapshots, which extensively boosts the automation assembly \cite{Ying2009, Yang2016a, Liu2016}. Thanks to 3D registration, the image stitching can be performed accurately for better sequence processing \cite{He2017}. And moreover, the navigation performance of intelligent vehicles can be improved by existing registration techniques \cite{Zhou2013,Zhou2016}. \\
\indent The basic 3D registration problem can usually be expressed as a least-square fitting one which takes the following form \cite{Kanatani1994}
\begin{equation}\label{opt}
\mathop {\arg \min }\limits_{\bm{C} \in SO(3), \bm{T} \in \mathbb{R}^3} {\mathcal{L}} = \sum\limits_{i = 1}^n {{a_i}{{\left\| {{{\bm{b}}_i} - {\bm{C}}{{\bm{r}}_i} - {\bm{T}}} \right\|}^2}} 
\end{equation}
where ${{{\bm{b}}_i}} \in \{{\cal{B}}\}$ and ${{{\bm{r}}_i}}  \in \{{\cal{R}}\}$ are point measurement pair in the body and reference frames; $a_i$ is the positive weight associated with the $i$-th point pair. The target is to estimate the direction cosine matrix ${\bm{C}}$ in the special orthogonal group $SO(3)$ such that ${\bm{C}}{{\bm{C}}^T} = {\bm{I}},\det ({\bm{C}}) =  + 1$, with ${\bm{T}}$ in the real 3D vector space $\mathbb{R}^3$, to minimize the sum $\mathcal{L}$. As there are both noise items inside ${{{\bm{b}}_i}}$ and ${{{\bm{r}}_i}}$, the problem is actually a total least square (TLS) \cite{Chang2015,Ruiter2014}. In real engineering applications, $\{\cal{B}\}$ and $\{\cal{R}\}$ do not always agree in the dimension. So the problem (\ref{opt}) is usually dealt with using the iterative closest points (ICP) for robust matching \cite{Besl1992}. Apart from ICP i.e. only find-based approach, the local geometric features inside the point clouds and images can also help for more robust matching \cite{Govindu1999}. Local geometric features are more advanced understanding and aspects inside data sequences and can reflect those regular and visualizable geometric characteristics \cite{Choi2009,Olson2001}, which intrinsically enhance the performance of 3D registration in urban areas e.g. in applications of the lidar odometry and mapping in real-time (LOAM, \cite{Zhang2017,Zhang2017a}). When there are rare local geometric features in the data, ICP is still generally feasible for processing global registration. Some algorithms have been proposed in the last 30 years to solve the $\bm{C}$ and $\bm{T}$ from (\ref{opt}) efficiently. The first famous solver was proposed by K. S. Arun et al. who introduces SVD for rotation estimation \cite{Arun1987}. However, when the problem contains large noise density, only SVD can not give robust estimation. Umeyama improves Arun's method by changing the signs of the singular values \cite{Umeyama1991}. In fact, the only difficulty of the optimization (\ref{opt}) is that $\bm{C}$ is nonlinear. However, after parameterizing $\bm{C}$ with dual quaternion, the problem can also be solved \cite{Walker1991}. A simpler approach is established by unit quaternions which converts the problem (\ref{opt}) into an EIG one \cite{Horn1987}. In fair comparisons, the dual quaternion is the slowest while EIG is slightly slower than SVD \cite{Pomerleau12comp}. But for both EIG and SVD, the numerical implementation requires many computation loads and space consumption of required libraries. This generates a difficulty for their usage on some critical platforms e.g. field programmable gate arrays (FPGA) and some low-configuration micro controller units (MCU) \cite{Guo2013}. The current situation also sets an obstacle for mass production of specified low-power integrated circuit (IC).\\
\indent Recently, we propose an algorithm for vector-observation attitude determination called the fast linear attitude estimator (FLAE) \cite{Wu2018}. FLAE owns the much superior computation speed compared with previous representatives. Motivated by SVD, EIG and FLAE, in this paper, a novel symbolic method is proposed. Through tests, the algorithm is verified to have only \%50 to \%80 execution time of recent fast SVD and EIG by C++ implementation on both personal computer and the MCU. In section II we present how to relate (\ref{opt}) with FLAE together. A simplified algorithm of eigenvalue is derived as well in this section. Experimental validations are presented in section III while section IV consists of concluding remarks. 

\section{Main Results}
The FLAE actually solves a more specific variant of (\ref{opt}) where $\bm{T}=0$ and $\left\| {{{\bm{b}}_i}} \right\| = \left\| {{{\bm{r}}_i}} \right\| = 1$ \cite{Wu2018} which can be further extended to some optimized results \cite{Wu2019tase}. This is a preliminary for attitude determination from normalized vector observations in spacecraft motion measurement. In FLAE, the DCM is parameterized by the unit quaternion. The optimal quaternion is associated with eigenvalue of $\bm{W}$ that is closest to 1 where $\bm{W}$ is given by
\begin{equation}\label{W}
\begin{array}{l}
{{\bm{W}}_{1,1}} = {H_{x1}} + {H_{y2}} + {H_{z3}}\\
{{\bm{W}}_{1,2}} =  - {H_{y3}} + {H_{z2}}\\
{{\bm{W}}_{1,3}} =  - {H_{z1}} + {H_{x3}}\\
{{\bm{W}}_{1,4}} =  - {H_{x2}} + {H_{y1}}\\
{{\bm{W}}_{2,1}} =  - {H_{y3}} + {H_{z2}}\\
{{\bm{W}}_{2,2}} = {H_{x1}} - {H_{y2}} - {H_{z3}}\\
{{\bm{W}}_{2,3}} = {H_{x2}} + {H_{y1}}\\
{{\bm{W}}_{2,4}} = {H_{x3}} + {H_{z1}}\\
{{\bm{W}}_{3,1}} =  - {H_{z1}} + {H_{x3}}\\
{{\bm{W}}_{3,2}} = {H_{x2}} + {H_{y1}}\\
{{\bm{W}}_{3,3}} = {H_{y2}} - {H_{x1}} - {H_{z3}}\\
{{\bm{W}}_{3,4}} = {H_{y3}} + {H_{z2}}\\
{{\bm{W}}_{4,1}} =  - {H_{x2}} + {H_{y1}}\\
{{\bm{W}}_{4,2}} = {H_{x3}} + {H_{z1}}\\
{{\bm{W}}_{4,3}} = {H_{y3}} + {H_{z2}}\\
{{\bm{W}}_{4,4}} = {H_{z3}} - {H_{y2}} - {H_{x1}}
\end{array}
\end{equation}
in which ${\bm{W}}_{i,j}$ denotes the matrix entry of $\bm{W}$ in the $i$-th row and $j$-th column. The parameters inside are provided as follows
\begin{equation}
{\bm{H}} = \left( {\begin{array}{*{20}{c}}
  {{H_{x1}}}&{{H_{y1}}}&{{H_{z1}}} \\ 
  {{H_{x2}}}&{{H_{y2}}}&{{H_{z2}}} \\ 
  {{H_{x3}}}&{{H_{y3}}}&{{H_{z3}}} 
\end{array}} \right) = \sum\limits_{i = 1}^n {{a_i}{{\bm{b}}_i}{\bm{r}}_i^T} 
\end{equation}
The characteristic polynomial of $\bm{W}$ takes the form of
\begin{equation} \label{chara}
{\lambda ^4} + {\tau _1}{\lambda ^2} + {\tau _2}\lambda  + {\tau _3}=0
\end{equation}
where
\begin{equation}\label{tau123}
\begin{array}{*{20}{l}}
{{\tau _1} = }
{ - 2\left( \begin{array}{l}
H_{x1}^2 + H_{x2}^2 + H_{x3}^2 + H_{y1}^2\\
 + H_{y2}^2 + H_{y3}^2 + H_{z1}^2 + H_{z2}^2 + H_{z3}^2
\end{array} \right)}\\
{{\tau _2} = }
{8({H_{x3}}{H_{y2}}{H_{z1}} - {H_{x2}}{H_{y3}}{H_{z1}} - {H_{x3}}{H_{y1}}{H_{z2}}}\\
\ \ \ \ \ \ \ \ \ \ \ \ \ \ \ \ { + {H_{x1}}{H_{y3}}{H_{z2}} + {H_{x2}}{H_{y1}}{H_{z3}} - {H_{x1}}{H_{y2}}{H_{z3}})}\\
{{\tau _3} = \det ({\bm{W}})}
\end{array}
\end{equation}
For the problem (\ref{opt}), the quaternion solution is produced by the optimal eigenvector of the following matrix \cite{Aghili2016, Besl1992}
\begin{equation}\label{G}
{\bm{G}} = \left[ {\begin{array}{*{20}{c}}
  {tr({\bm{D}})}&{{{\bm{z}}^T}} \\ 
  {\bm{z}}&{{\bm{D}} + {{\bm{D}}^T} - tr({\bm{D}}){\bm{I}}} 
\end{array}} \right]
\end{equation}
in which
\begin{equation}
\begin{gathered}
  {\bm{D}} = \sum\limits_{i = 1}^n {{a_i}({{\bm{b}}_i} - {\bm{\bar b}}){{({\bm{r}}_i^{} - {\bm{\bar r}})}^T}}  \hfill \\
  {\bm{z}} = \sum\limits_{i = 1}^n {{a_i}({{\bm{b}}_i} - {\bm{\bar b}}) \times ({\bm{r}}_i^{} - {\bm{\bar r}})}  \hfill \\
  {\bm{\bar b}} = \sum\limits_{i = 1}^n a_i{{{\bm{b}}_i}} ,{\bm{\bar r}} = \sum\limits_{i = 1}^n a_i{{{\bm{r}}_i}}  \hfill \\ 
\end{gathered} 
\end{equation}
It is obvious that $\bm{D}$ has the same structure with $\bm{H}$. Then if $\bm{D}=H$, we would like to prove that $\bm{G}=W$. It can be directly obtained that
\begin{equation}
\begin{gathered}
  {\bm{D}} + {{\bm{D}}^T} - tr({\bm{D}}){\bm{I}} =  \hfill \\
\begin{small}  \left( {\begin{array}{*{20}{c}}
  {{H_{x1}} - {H_{y2}} - {H_{z3}}}&{{H_{x2}} + {H_{y1}}}&{{H_{x3}} + {H_{z1}}} \\ 
  {{H_{x2}} + {H_{y1}}}&{{H_{y2}} - {H_{x1}} - {H_{z3}}}&{{H_{y3}} + {H_{z2}}} \\ 
  {{H_{x3}} + {H_{z1}}}&{{H_{y3}} + {H_{z2}}}&{{H_{z3}} - {H_{y2}} - {H_{x1}}} 
\end{array}} \right) \end{small} \hfill \\
  tr({\bm{D}}) = {H_{x1}} + {H_{y2}} + {H_{z3}} \hfill \\
\end{gathered}
\end{equation}
For $\bm{z}$, it has another form according to the skew-symmetric matrix of cross-product, such that
\begin{equation}
\begin{gathered}
  {\bm{z}} = {\left( {{{\bm{D}}_{2,3}} - {{\bm{D}}_{3,2}},{{\bm{D}}_{3,1}} - {{\bm{D}}_{1,3}},{{\bm{D}}_{1,2}} - {{\bm{D}}_{2,1}}} \right)^T} \hfill \\
  \begin{array}{*{20}{c}}
  {}&{} 
\end{array} = {\left( { - {H_{y3}} + {H_{z2}}, - {H_{z1}} + {H_{x3}}, - {H_{x2}} + {H_{y1}}} \right)^T} \hfill \\ 
\end{gathered} 
\end{equation}
Inserting these results into (\ref{G}), one can observe that $\bm{W} = G$. So the characteristic polynomial of $\bm{W}$ can also be used for eigenvalue solving of $\bm{G}$.\\
\indent The FLAE gives the following symbolic roots of (\ref{chara}):
\begin{equation}
\begin{array}{l}
{\lambda _1} = \frac{1}{{2\sqrt 6 }}\left( {{T_2} - \sqrt { - T_2^2 - 12{\tau _1} - \frac{{12\sqrt 6 {\tau _2}}}{{{T_2}}}} } \right)\\
{\lambda _2} = \frac{1}{{2\sqrt 6 }}\left( {{T_2} + \sqrt { - T_2^2 - 12{\tau _1} - \frac{{12\sqrt 6 {\tau _2}}}{{{T_2}}}} } \right)\\
{\lambda _3} =  - \frac{1}{{2\sqrt 6 }}\left( {{T_2} + \sqrt { - T_2^2 - 12{\tau _1} + \frac{{12\sqrt 6 {\tau _2}}}{{{T_2}}}} } \right)\\
{\lambda _4} =  - \frac{1}{{2\sqrt 6 }}\left( {{T_2} - \sqrt { - T_2^2 - 12{\tau _1} + \frac{{12\sqrt 6 {\tau _2}}}{{{T_2}}}} } \right)
\end{array}
\end{equation}
in which
\begin{equation}
\begin{array}{l}
{T_0} = 2\tau _1^3 + 27\tau _2^2 - 72\tau _1^{}\tau _3^{}\\
{T_1} = {\left( {{T_0} + \sqrt { - 4{{\left( {\tau _1^2 + 12\tau _3^{}} \right)}^3} + T_0^2} } \right)^{\frac{1}{3}}}\\
{T_2} = \sqrt { - 4\tau _1^{} + \frac{{{2^{\frac{4}{3}}}\left( {\tau _1^2 + 12\tau _3^{}} \right)}}{{{T_1}}} + {2^{\frac{2}{3}}}{T_1}} 
\end{array}
\end{equation}
Let us first determine the signs of $\tau_1, \tau_2, \tau_3$. $\bm{W}$ is real symmetric and the eigenvalues are two positive and two negative. This gives $\tau_3=\det({\bm{W}})=\lambda_1 \lambda_2 \lambda_3 \lambda_4 > 0$. $\tau_1$ is obvious negative and $\tau_2$ is indefinite. In this way, $T_0$ is definitely real number. Let us write $T_1, T_2$ into
\begin{equation}
\begin{gathered}
  {T_1} = {\alpha _{{T_1}}} + {\beta _{{T_1}}}{\bm{i}} \hfill \\
  {T_2} = {\alpha _{{T_2}}} + {\beta _{{T_2}}}{\bm{i}} \hfill \\ 
\end{gathered} 
\end{equation}
where $\bm{i}$ denotes the unit imaginary number while ${\alpha _{{T_1}}},{\beta _{{T_1}}},{\alpha _{{T_2}}},{\beta _{{T_2}}} \in \mathbb{R}$. Obviously, $T_1$ meets
\begin{equation}
\begin{gathered}
  T_1^3 = \alpha _{{T_1}}^3 - 3{\alpha _{{T_1}}}\beta _{{T_1}}^2 + \left( {3\alpha _{{T_1}}^2{\beta _{{T_1}}} - \beta _{{T_1}}^3} \right){\bm{i}} \hfill \\
   = {T_0} + \sqrt { - 4{{\left( {\tau _1^2 + 12{\tau _3}} \right)}^3} + T_0^2}  \hfill \\ 
\end{gathered} 
\end{equation}
Likewise, we have
\begin{equation}
\begin{gathered}
  T_2^2 = \alpha _{{T_2}}^2 - \beta _{{T_2}}^2 + 2{\alpha _{{T_2}}}{\beta _{{T_2}}}{\bm{i}} \hfill \\
   =  - 4{\tau _1} + \frac{{{2^{\frac{4}{3}}}\left( {\tau _1^2 + 12{\tau _3}} \right)}}{{{T_1}}} + {2^{\frac{2}{3}}}{T_1} \hfill \\
   =  - 4{\tau _1} + \frac{{{2^{\frac{4}{3}}}\left( {\tau _1^2 + 12{\tau _3}} \right)}}{{{\alpha _{{T_1}}} + {\beta _{{T_1}}}{\bm{i}}}} + {2^{\frac{2}{3}}}\left( {{\alpha _{{T_1}}} + {\beta _{{T_1}}}{\bm{i}}} \right) \hfill \\
   =  - 4{\tau _1} + \left[ {\sqrt[3]{4} + \frac{2{\sqrt[3]{2}\left( {\tau _1^2 + 12{\tau _3}} \right)}}{{\alpha _{{T_1}}^2 + \beta _{{T_1}}^2}}} \right]{\alpha _{{T_1}}} \hfill \\
  \begin{array}{*{20}{c}}
  {}&{} 
\end{array} + \left[ {\sqrt[3]{4} - \frac{2{\sqrt[3]{2}\left( {\tau _1^2 + 12{\tau _3}} \right)}}{{\alpha _{{T_1}}^2 + \beta _{{T_1}}^2}}} \right]{\beta _{{T_1}}}{\bm{i}} \hfill \\ 
\end{gathered} 
\end{equation}
These equations lead to the system of
\begin{equation}\label{system}
\left\{ {\begin{array}{*{20}{c}}
  {\alpha _{{T_1}}^3 - 3{\alpha _{{T_1}}}\beta _{{T_1}}^2 = {T_0}} \\ 
  {{{\left( {3\alpha _{{T_1}}^2{\beta _{{T_1}}} - \beta _{{T_1}}^3} \right)}^2} = 4{{\left( {\tau _1^2 + 12{\tau _3}} \right)}^3} - T_0^2} \\ 
  {\alpha _{{T_2}}^2 - \beta _{{T_2}}^2 =  - 4{\tau _1} + \left[ {\sqrt[3]{4} + \frac{{2\sqrt[3]{2}\left( {\tau _1^2 + 12{\tau _3}} \right)}}{{\alpha _{{T_1}}^2 + \beta _{{T_1}}^2}}} \right]{\alpha _{{T_1}}}} \\ 
  {2{\alpha _{{T_2}}}{\beta _{{T_2}}} = \left[ {\sqrt[3]{4} - \frac{{2\sqrt[3]{2}\left( {\tau _1^2 + 12{\tau _3}} \right)}}{{\alpha _{{T_1}}^2 + \beta _{{T_1}}^2}}} \right]{\beta _{{T_1}}}} 
\end{array}} \right.
\end{equation}
From the first two sub-equations, one can easily arrive at
\begin{equation}\label{equal}
\begin{gathered}
  {\left( {\alpha _{{T_1}}^3 - 3{\alpha _{{T_1}}}\beta _{{T_1}}^2} \right)^2} + {\left( {3\alpha _{{T_1}}^2{\beta _{{T_1}}} - \beta _{{T_1}}^3} \right)^2} = 4{\left( {\tau _1^2 + 12{\tau _3}} \right)^3} \hfill \\
   \Rightarrow \alpha _{{T_1}}^6 + 3\alpha _{{T_1}}^4\beta _{{T_1}}^2 + 3\alpha _{{T_1}}^2\beta _{{T_1}}^4 + \beta _{{T_1}}^6 = 4{\left( {\tau _1^2 + 12{\tau _3}} \right)^3} \hfill \\
   \Rightarrow \alpha _{{T_1}}^2 + \beta _{{T_1}}^2 = \sqrt[3]{4}\left( {\tau _1^2 + 12{\tau _3}} \right) \hfill \\ 
\end{gathered} 
\end{equation}
Inserting (\ref{equal}) into (\ref{system}) we have
\begin{equation}
\left\{ {\begin{array}{*{20}{c}}
  {\alpha _{{T_2}}^2 - \beta _{{T_2}}^2 =  - 4{\tau _1} + 2\sqrt[3]{4}{\alpha _{{T_1}}}} \\ 
  {2{\alpha _{{T_2}}}{\beta _{{T_2}}} = 0} 
\end{array}} \right.
\end{equation}
This indicates that ${\alpha _{{T_2}}}=0$ or ${\beta _{{T_2}}}=0$. If ${\alpha _{{T_2}}}=0$ then $T_2$ is a pure imaginary number leading to the eigenvalues of complex numbers, which is not true for real symmetric matrix. Therefore we have ${\beta _{{T_2}}}=0$ i.e. $T_2$ is a pure positive real number with no imaginary part. Using this finding, the maximum eigenvalue is immediately $\lambda_2$. The components of $T_1$ can be computed using
\begin{equation}\label{tmp}
\begin{gathered}
  T_1^3 = {T_0} + \sqrt { - 4{{\left( {\tau _1^2 + 12{\tau _3}} \right)}^3} + T_0^2}  \hfill \\
   = {T_0} + \sqrt {4{{\left( {\tau _1^2 + 12{\tau _3}} \right)}^3} - T_0^2} {\bm{i}} \hfill \\
   = 2{\left( {\tau _1^2 + 12{\tau _3}} \right)^{\frac{3}{2}}}{e^{{\bm{i}}\arctan \frac{{\sqrt {4{{\left( {\tau _1^2 + 12{\tau _3}} \right)}^3} - T_0^2} }}{{{T_0}}}}} \hfill \\ 
\end{gathered} 
\end{equation}
After the maximum eigenvalue is computed, the elementary row operations are needed to calculate the associated eigenvector from $\left( {{\bm{G}} - {\lambda _{\max }}{\bm{I}}} \right){\bm{q}} = {\bm{0}}$. Given an arbitrary real symmetric matrix below \cite{Wu2018}
\begin{equation}\label{GG}
{\bm{G}} - {\lambda _{\max }}{\bm{I}} = \left( {\begin{array}{*{20}{c}}
  {{G_{11}}}&{{G_{12}}}&{{G_{13}}}&{{G_{14}}} \\ 
  {{G_{12}}}&{{G_{22}}}&{{G_{23}}}&{{G_{24}}} \\ 
  {{G_{13}}}&{{G_{23}}}&{{G_{33}}}&{{G_{34}}} \\ 
  {{G_{14}}}&{{G_{24}}}&{{G_{34}}}&{{G_{44}}} 
\end{array}} \right)
\end{equation}
The optimal quaternion $\bm{q}$ from row operations can be categorized as follows
\begin{equation}\label{q}
\begin{small}
\begin{gathered}
  {q_0} = {G_{14}}G_{23}^2 - {G_{13}}{G_{24}}{G_{23}} - {G_{12}}{G_{34}}{G_{23}} -  \hfill \\
  \begin{array}{*{20}{c}}
  {}&{} 
\end{array}{G_{14}}{G_{22}}{G_{33}} + {G_{12}}{G_{24}}{G_{33}} + {G_{13}}{G_{22}}{G_{34}} \hfill \\
  {q_1} = {G_{24}}G_{13}^2 - {G_{12}}{G_{34}}{G_{13}} - {G_{13}}{G_{14}}{G_{23}} +  \hfill \\
  \begin{array}{*{20}{c}}
  {}&{} 
\end{array}{G_{12}}{G_{14}}{G_{33}} - {G_{11}}{G_{24}}{G_{33}} + {G_{11}}{G_{23}}{G_{34}} \hfill \\
  {q_2} = {G_{34}}G_{12}^2 - {G_{14}}{G_{23}}{G_{12}} - {G_{13}}{G_{24}}{G_{12}} +  \hfill \\
  \begin{array}{*{20}{c}}
  {}&{} 
\end{array}{G_{13}}{G_{14}}{G_{22}} + {G_{11}}{G_{23}}{G_{24}} - {G_{11}}{G_{22}}{G_{34}} \hfill \\
  {q_3} =  - {G_{33}}G_{12}^2 + 2{G_{13}}{G_{23}}{G_{12}} - {G_{11}}G_{23}^2 - G_{13}^2{G_{22}} \hfill \\
  \begin{array}{*{20}{c}}
  {}&{} 
\end{array} + {G_{11}}{G_{22}}{G_{33}} \hfill \\ 
\end{gathered} 
\end{small}
\end{equation}
where ${\bm{q}} = {\left( {{q_0},{q_1},{q_2},{q_3}} \right)^T}$. The estimated attitude quaternion is then ${\bm{\hat q}} = {\bm{q}}/\left\| {\bm{q}} \right\|$.

\begin{table*}[hb]
\centering
\caption{Studied Cases for Comparisons}\label{tab:cases}
\resizebox{1.0\textwidth}{!}{
\begin{tabular}{cccccc}
\toprule
{Case}&{Euler Angles $\varphi ,\vartheta ,\psi $}&{Translation $\bm{T}$}&{Noise Covariance ${{\bm{\Sigma}} _{{\bm{\eta} _i}}}$}&{Vector Number $n$}&{${\rm{rank}}({\bm{D}})$}\\
\midrule
{1}&{$\left( { - \frac{\pi }{6},\frac{{4\pi }}{{11}}, - \frac{{5\pi }}{7}} \right)=(-0.52359878, 1.1423973, -2.2439948)$}&{${\left( {100, - 50,80} \right)^T}$}&{$diag\left( {0.0,0.0,0.0} \right)$}&{100}&{3}\\
{2}&{$\left( { - \frac{\pi }{6},\frac{{4\pi }}{{11}}, - \frac{{5\pi }}{7}} \right)=(-0.52359878, 1.1423973, -2.2439948)$}&{${\left( {100, - 50,80} \right)^T}$}&{$diag\left( {0.0,0.0,0.0} \right)$}&{100}&{2}\\
{3}&{$\left( { - \frac{\pi }{6},\frac{{4\pi }}{{11}}, - \frac{{5\pi }}{7}} \right)=(-0.52359878, 1.1423973, -2.2439948)$}&{${\left( {100, - 50,80} \right)^T}$}&{$diag\left( {0.0,0.0,0.0} \right)$}&{100}&{1}\\
{4}&{$\left( { \frac{4 \pi }{7},\frac{{\pi }}{{2}}, - \frac{{9\pi }}{20}} \right)=(1.7951958, 1.5707963, -1.4137167)$}&{${\left( {-60, 70, 40} \right)^T}$}&{$diag\left( {10,10,10} \right)$}&{100}&{3}\\
{5}&{$\left( { \frac{4 \pi }{7},\frac{{\pi }}{{2}}, - \frac{{9\pi }}{20}} \right)=(1.7951958, 1.5707963, -1.4137167)$}&{${\left( {-60, 70, 40} \right)^T}$}&{$diag\left( {10,10,10} \right)$}&{1000}&{3}\\
{6}&{$\left( { \frac{4 \pi }{7},\frac{{\pi }}{{2}}, - \frac{{9\pi }}{20}} \right)=(1.7951958, 1.5707963, -1.4137167)$}&{${\left( {-60, 70, 40} \right)^T}$}&{$diag\left( {10,10,10} \right)$}&{10000}&{3}\\
{7}&{$\left( { \frac{5 \pi }{9}, - \frac{{7\pi }}{{10}}, \frac{{4\pi }}{13}} \right)=(-1.3962634, -0.9424778, -2.1749488)$}&{${\left( {80, -20, -160} \right)^T}$}&{$diag\left( {0.1,10,1000} \right)$}&{1000}&{3}\\
{8}&{$\left( { \frac{5 \pi }{9}, - \frac{{7\pi }}{{10}}, \frac{{4\pi }}{13}} \right)=(-1.3962634, -0.9424778, -2.1749488)$}&{${\left( {80, -20, -160} \right)^T}$}&{$diag\left( {1000,10,0.1} \right)$}&{1000}&{3}\\
{9}&{$\left( { \frac{5 \pi }{9}, - \frac{{7\pi }}{{10}}, \frac{{4\pi }}{13}} \right)=(-1.3962634, -0.9424778, -2.1749488)$}&{${\left( {80, -20, -160} \right)^T}$}&{$diag\left( {0.1,0,1,0.1} \right)$}&{1000}&{3}\\
\bottomrule
\\
\bottomrule
\end{tabular}}
\end{table*}

\subsection{Numerical Robustness}
Here the numerical robustness of the proposed method is referred to the behavior when the two largest eigenvalues almost coincide. In such extreme case, the measurements from $\{ {\cal{B}} \}$ and $\{ {\cal{R}} \}$ are basically collinear \cite{Cheng2014}. This makes $\bm{D}$ almost a rank-deficient matrix. Then, we immediately have
 \begin{equation}
 {\tau _2} =  - 8\det \left( {\bm{D}} \right) \approx 0
 \end{equation}
  and also
 \begin{equation}
 \left\{ {\begin{array}{*{20}{c}}
  {\tau _1^2 - 4{\tau _3} = 0} \\ 
  {{\tau _3} = \det ({\bm{W}}) = {\lambda _1}{\lambda _2}{\lambda _3}{\lambda _4} = \lambda _{\max }^4} 
\end{array}} \right.
 \end{equation}
 Inserting $\tau _1^2 = 4{\tau _3}$ into (\ref{tmp}), it is quite straightforward for one to obtain 
 \begin{equation}
\begin{gathered}
  \sqrt {4{{\left( {\tau _1^2 + 12{\tau _3}} \right)}^3} - T_0^2}  \approx 0 \hfill \\
   \Rightarrow \left\{ {\begin{array}{*{20}{c}}
  {\theta  \approx 0} \\ 
  {{\alpha _{{T_1}}} \approx  - 2\sqrt[3]{2}{\tau _1}} 
\end{array}} \right. \hfill 
   \Rightarrow {T_2} \approx 0 \hfill \\ 
\end{gathered} 
 \end{equation}
  Note that here, $T_2$ and $\tau_2$ both approach 0 and there is an indefinite limit in the eigenvalue i.e. $\mathop {\lim }\limits_{{\tau _2} \to 0} \frac{{{\tau _2}}}{{{T_2}}}$. Repeating the L'Hospital rule, we can eventually arrive at
 \begin{equation}
\begin{small}
\begin{gathered}
  \mathop {\lim }\limits_{{\tau _2} \to 0} \frac{{{\tau _2}}}{{{T_2}}} \hfill \\
   = \mathop {\lim }\limits_{{\tau _2} \to 0} \begin{small} \frac{{d{\tau _2}}}{{d\sqrt {\begin{gathered}
   - 4\tau _1^{} +  \hfill \\
  \frac{{{2^{\frac{4}{3}}}\left( {\tau _1^2 + 12\tau _3^{}} \right)}}{{{{\left( \begin{gathered}
  2\tau _1^3 + 27\tau _2^2 - 72\tau _1^{}\tau _3^{} +  \hfill \\
  \sqrt { - 4{{\left( {\tau _1^2 + 12\tau _3^{}} \right)}^3} + {{\left( {2\tau _1^3 + 27\tau _2^2 - 72\tau _1^{}\tau _3^{}} \right)}^2}}  \hfill \\ 
\end{gathered}  \right)}^{\frac{1}{3}}}}} \hfill \\
   + {2^{\frac{2}{3}}}{\left( \begin{gathered}
  2\tau _1^3 + 27\tau _2^2 - 72\tau _1^{}\tau _3^{} +  \hfill \\
  \sqrt { - 4{{\left( {\tau _1^2 + 12\tau _3^{}} \right)}^3} + {{\left( {2\tau _1^3 + 27\tau _2^2 - 72\tau _1^{}\tau _3^{}} \right)}^2}}  \hfill \\ 
\end{gathered}  \right)^{\frac{1}{3}}} \hfill \\ 
\end{gathered}}  }} \end{small} \hfill \\
   = 0 \hfill \\ 
\end{gathered} 
\end{small}
 \end{equation}
 where $d$ is the differentiation operator. Therefore, the limiting maximum eigenvalue is
 \begin{equation}\label{lambda_ex}
 {\lambda _{\max }} \approx \sqrt { - \frac{{{\tau _1}}}{2}} 
 \end{equation}
 This indicates that in extreme cases, the eigenvalue is still not singular which always leads to meaningful quaternion solutions. However, for iterative algorithms like Gauss-Newton iteration, the solving process can hardly stop according to word length of floating numbers \cite{Cheng2014}. This shows that the proposed method may be more practical in real engineering implementation.
The final computation procedure is summarized in Algorithm 1.

\begin{algorithm}[H] 
\label{algorithm:FS3R} 
\caption{The Fast Symbolic 3D Registration (FS3R) Algorithm} 
\textbf{Require:} Point correspondences $\{ {\cal{B}} \}$ and $\{ {\cal{R}} \}$ with same dimension of $n$, provided that the weights $\{ a_i, i=1,2,3,\cdots \}$ exist. If no weights, each weight is equalized to $\frac{1}{a}$. The numerical tolerance threshold for detecting extreme case is defined as $\xi$ which is normally a very tiny positive number.\\
\textbf{Step 1:} Calculate mean points ${\bm{\bar b}} = \sum\limits_{i = 1}^n {{a_i}{{\bm{b}}_i}} ,{\bm{\bar r}} = \sum\limits_{i = 1}^n {{a_i}{{\bm{r}}_i}} $.\\
\textbf{Step 2:} Compute $\bm{H}$ matrix using simplified form \\${\bm{H}} = \sum\limits_{i = 1}^n {{a_i}({{\bm{b}}_i}{\bm{r}}_i^T - {\bm{\bar b}}{{{\bm{\bar r}}}^T})} $ and then compute $\bm{W}$ using (\ref{W}).\\
\textbf{Step 3:} Compute coefficients of characteristic polynomial \\from (\ref{tau123}).\\
\textbf{Step 4:} Compute ${T_0} = 2\tau _1^3 + 27\tau _2^2 - 72\tau _1^{}\tau _3^{}$ and then compute $T_1$ by 
$\begin{small} \begin{gathered}
  \theta  = \arctan \frac{{\sqrt {4{{\left( {\tau _1^2 + 12{\tau _3}} \right)}^3} - T_0^2} }}{{{T_0}}} \hfill \\
  {\alpha _{{T_1}}} = \sqrt[3]{2}\sqrt {\tau _1^2 + 12{\tau _3}} \cos \frac{\theta }{3} \hfill \\
  {\beta _{{T_1}}} = \sqrt[3]{2}\sqrt {\tau _1^2 + 12{\tau _3}} \sin \frac{\theta }{3} \hfill \\ 
\end{gathered} \end{small}$\\
\textbf{Step 5:} Compute\\
${T_2} = \left| {{\alpha _{{T_2}}}} \right| = \sqrt { - 4{\tau _1} + 2\sqrt[3]{4}{\alpha _{{T_1}}}} $. If $\left| {{\tau _2}} \right| > \xi ,\left| {{T_2}} \right| > \xi $, then compute the maximum eigenvalue ${\lambda _{\max }} = \frac{1}{{2\sqrt 6 }}\left( {{T_2} + \sqrt { - T_2^2 - 12{\tau _1} - \frac{{12\sqrt 6 {\tau _2}}}{{{T_2}}}} } \right)$. Else, compute eigenvalue according to (\ref{lambda_ex}).\\
\
\textbf{Step 6:} Compute required elements in (\ref{GG}) and then calculate the normalized unit quaternion according to (\ref{q}).\\
\textbf{Step 7:} Reconstruct the rotation from quaternion as $\bm{C}$. The translation is computed by ${\bm{T}} = {\bm{\bar b}} - {\bm{C\bar r}}$.
\label{algorithm:FS3R} 
\end{algorithm}

\begin{table*}[ht]
\centering
\caption{Estimated Euler Angles $\varphi ,\vartheta ,\psi $}\label{tab:euler}
\resizebox{1.0\textwidth}{!}{
\begin{tabular}{cccccc}
\toprule
{Case}&{SVD}&{EIG}&{EIG Analytical}&{SVD Analytical}&{Proposed FS3R}\\
\midrule
{1}&{$(-0.5235, 1.1424, -2.2439)$}&{$(-0.5235, 1.1424, -2.2439)$}&{$(-0.5235, 1.1424, -2.2439)$}&{$(-0.5235, 1.1424, -2.2439)$}&{$\mathbf{(-0.5235, 1.1424, -2.2439)}$}\\
{2}&{$(-2.8874, 0.6156, -2.3558)$}&{$(1.3088, 0.6156, -2.3558)$}&{$(1.3088, 0.6156, -2.3558)$}&{$(NaN, NaN, NaN)$}&{$\mathbf{(1.3088, 0.6156, -2.3558)}$}\\
{3}&{$(-0.04696, -0.04481, -0.04696)$}&{$(-2.0344, 0.7297, -2.0344)$}&{$(-2.0344, 0.7297, -2.0344)$}&{$(0.4621, - 1.571 + 10.693{\bm{i}}, 2.356)$}&{$\mathbf{(-2.0344, 0.7297, -2.0344)}$}\\
{4}&{$(0.3614, 1.258, -0.5719)$}&{$(0.3614, 1.258, -0.5719)$}&{$(0.3614, 1.258, -0.5719)$}&{$(0.3614, 1.258, -0.5719)$}&{$\mathbf{(0.3614, 1.258, -0.5719)}$}\\
{5}&{$(0.6665, 1.4052, -0.4474)$}&{$(0.6665, 1.4052, -0.4474)$}&{$(0.6665, 1.4052, -0.4474)$}&{$(0.6665, 1.4052, -0.4474)$}&{$\mathbf{(0.6665, 1.4052, -0.4474)}$}\\
{6}&{$(-0.02599, 1.4942, 0.4477)$}&{$(-0.02599, 1.4942, 0.4477)$}&{$(-0.02599, 1.4942, 0.4477)$}&{$(-0.02599, 1.4942, 0.4477)$}&{$\mathbf{(-0.02599, 1.4942, 0.4477)}$}\\
{7}&{$(2.7465, 0.5139, 2.7899)$}&{$(2.7465, 0.5139, 2.7899)$}&{$(2.7465, 0.5139, 2.7899)$}&{$(2.7465, 0.5139, 2.7899)$}&{$\mathbf{(2.7465, 0.5139, 2.7899)}$}\\
{8}&{$(0.2577, -0.3486, 0.2181)$}&{$(0.2577, -0.3486, 0.2181)$}&{$(0.2577, -0.3486, 0.2181)$}&{$(0.2577, -0.3486, 0.2181)$}&{$\mathbf{(0.2577, -0.3486, 0.2181)}$}\\
{9}&{$(-1.4018, -0.9443, -2.1804)$}&{$(-1.4018, -0.9443, -2.1804)$}&{$(-1.4018, -0.9443, -2.1804)$}&{$(-1.4018, -0.9443, -2.1804)$}&{$\mathbf{(-1.4018, -0.9443, -2.1804)}$}\\
\bottomrule
\bottomrule
\ \\
\end{tabular}}

\centering
\caption{Estimated Translation $\bm{T}$}\label{tab:trans}
\resizebox{1.0\textwidth}{!}{
\begin{tabular}{cccccc}
\toprule
{Case}&{SVD}&{EIG}&{EIG Analytical}&{SVD Analytical}&{Proposed FS3R}\\
\midrule
{1}&{$(100.015, -50.0834, 79.9858)^T$}&{$(100.015, -50.0834, 79.9858)^T$}&{$(100.015, -50.0834, 79.9858)^T$}&{$(100.015, -50.0834, 79.9858)^T$}&{$\mathbf{(100.015, -50.0834, 79.9858)^T}$}\\
{2}&{$(99.8532, -49.9929, -49.9929)^T$}&{$(99.8532, -49.9929, -49.9929)^T$}&{$(99.8532, -49.9929, -49.9929)^T$}&{$(NaN,NaN,NaN)^T$}&{$\mathbf{(99.8532, -49.9929, -49.9929)^T}$}\\
{3}&{$(100.0, 100.0, 100.0)^T$}&{$(100.0, 100.0, 100.0)^T$}&{$(100.0, 100.0, 100.0)^T$}&{$(100.354, 100.354, 100.354)^T$}&{$\mathbf{(100.0, 100.0, 100.0)^T}$}\\
{4}&{$(-59.3406, 69.5444, 39.2757)^T$}&{$(-59.3406, 69.5444, 39.2757)^T$}&{$(-59.3406, 69.5444, 39.2757)^T$}&{$(-59.3406, 69.5444, 39.2757)^T$}&{$\mathbf{(-59.3406, 69.5444, 39.2757)^T}$}\\
{5}&{$(-59.8461, 69.6513, 40.1395)^T$}&{$(-59.8461, 69.6513, 40.1395)^T$}&{$(-59.8461, 69.6513, 40.1395)^T$}&{$(-59.8461, 69.6513, 40.1395)^T$}&{$\mathbf{(-59.8461, 69.6513, 40.1395)^T}$}\\
{6}&{$(-59.8461, 69.6513, 40.1395)^T$}&{$(-59.8461, 69.6513, 40.1395)^T$}&{$(-59.8461, 69.6513, 40.1395)^T$}&{$(-59.8461, 69.6513, 40.1395)^T$}&{$\mathbf{(-59.8461, 69.6513, 40.1395)^T}$}\\
{7}&{$(79.9458, -19.9293, 141.043)^T$}&{$(79.9458, -19.9293, 141.043)^T$}&{$(79.9458, -19.9293, 141.043)^T$}&{$(79.9458, -19.9293, 141.043)^T$}&{$\mathbf{(79.9458, -19.9293, 141.043)^T}$}\\
{8}&{$(91.9475, -20.049, 160.038)^T$}&{$(91.9475, -20.049, 160.038)^T$}&{$(91.9475, -20.049, 160.038)^T$}&{$(91.9475, -20.049, 160.038)^T$}&{$\mathbf{(91.9475, -20.049, 160.038)^T}$}\\
{9}&{$(79.9251, -20.0097, 159.997)^T$}&{$(79.9251, -20.0097, 159.997)^T$}&{$(79.9251, -20.0097, 159.997)^T$}&{$(79.9251, -20.0097, 159.997)^T$}&{$\mathbf{(79.9251, -20.0097, 159.997)^T}$}\\
\bottomrule
\bottomrule
\ \\
\end{tabular}}

\centering
\caption{Loss Function Value $\mathcal{L}$ in (\ref{opt})}\label{tab:metric}
\resizebox{0.55\textwidth}{!}{
\begin{tabular}{cccccc}
\toprule
{Case}&{SVD}&{EIG}&{EIG Analytical}&{SVD Analytical}&{Proposed FS3R}\\
\midrule
{1}&{$0.01440$}&{$0.01440$}&{$0.01440$}&{$0.01440$}&{$\mathbf{0.01440}$}\\
{2}&{$0.24102$}&{$0.24102$}&{$0.24102$}&{$NaN$}&{$\mathbf{0.24102}$}\\
{3}&{$0.00222$}&{$0.00222$}&{$0.00222$}&{$0.00489$}&{$\mathbf{0.00222}$}\\
{4}&{$284.54905$}&{$284.54905$}&{$284.54905$}&{$284.54905$}&{$\mathbf{284.54905}$}\\
{5}&{$302.03084$}&{$302.03084$}&{$302.03084$}&{$302.03084$}&{$\mathbf{302.03084}$}\\
{6}&{$298.76512$}&{$298.76512$}&{$298.76512$}&{$298.76512$}&{$\mathbf{298.76512}$}\\
{7}&{$966940.84856$}&{$966940.84856$}&{$966940.84856$}&{$966940.84856$}&{$\mathbf{966940.84856}$}\\
{8}&{$977583.31035$}&{$977583.31035$}&{$977583.31035$}&{$977583.31035$}&{$\mathbf{977583.31035}$}\\
{9}&{$0.03313$}&{$0.03313$}&{$0.03313$}&{$0.03313$}&{$\mathbf{0.03313}$}\\
\bottomrule
\end{tabular}}
\end{table*}

\section{Experiments and Comparisons}
In this section, several experiments are conducted to present comparisons of the proposed fast symbolic 3D registration (FS3R) algorithm with representatives. Note that recently, some similar analytical methods have been proposed. For instance, Yang et al. developed an analytical method for root-solving of quartic equation \cite{Yang2013}. And a novel analytical SVD method is proposed recently by us to conduct factorization of $3\times 3$ matrix \cite{Liu2018}. These methods are faster than representative numerical ones. Therefore we mainly compare them with the proposed FS3R on the accuracy, robustness and computation speed. The algorithms are first implemented using MATLAB for validation of accuracy and robustness. They are then translated into C++ programming language for rigorous execution time performance test on both PC and ARM processors.

\subsection{Accuracy and Robustness Performance}
In this sub-section, the statistics are collected using the MATLAB r2016a software on a MacBook Pro 2017 with CPU clock speed of i7 4-core 3.5GHz. Here, simulated samples with different dimensions and noise density are generated by means of
\begin{equation}
{{\bm{b}}_i} = {\bm{C}}{{\bm{r}}_i} + {\bm{T}} + {\bm{\eta} _i},i = 1,2, \cdots ,n
\end{equation}
where $\bm{\eta}_i$ denotes the noise item subject to normal distribution with zero mean and covariance of ${{\bm{\Sigma}} _{{\bm{\eta} _i}}}$. By designing the experiments in Table \ref{tab:cases}, we evaluate the accuracy and robustness performance of various algorithms. The first cases employ the same rotation and translation while they differ mainly in ${\rm{rank}}(\bm{D})$ column. When ${\rm{rank}}(\bm{D}) < 3$, the case is defined to be extreme and some methods will fail to converge. Cases $4\sim 6$ consist of comparisons with different vector numbers. In cases $7 \sim 9$, we mainly describe the effect of the noise density. The evaluated results are shown in the Table \ref{tab:euler}, \ref{tab:trans} and \ref{tab:metric} for rotation, translation and loss function value $\mathcal{L}$ in (\ref{opt}), respectively. The rotation matrix is first estimated and then converted to the Euler angles i.e. roll $\varphi$, pitch $\vartheta$ and yaw $\psi$ through '$X-Y-Z$' rotation sequence. The $NaN$ value stands for the 'Not a Number' one which is usually caused by indefinite devisions like $\frac{0}{0}$ and $\frac{\infty}{\infty }$. Here, the 'SVD' and 'EIG' are implemented using MATLAB internal functions while 'EIG Analytical' is from \cite{Yang2013} and 'SVD Analytical' refers to \cite{Liu2018}.\\
\indent From the computed results, one can immediately observe from cases 1 to 3 that the robustness of the proposed FS3R maintains the same level with 'SVD', 'EIG' and 'EIG Analytical'. While in all these statistics, 'SVD Analytical' is the most weak one due to its low immunity to matrix rank deficiency. In the computation procedure, some steps break according to numerical problems and thus generate $NaN$ values. Such disadvantage is deadly because once this happens in an embedded computation system, without proper detection, the system is very likely to crash since these digits are meaningless. The cases $4\sim 9$ describes general accuracy of various algorithms. From cases $4$ to $6$, the number of vectors increases. Then from Table \ref{tab:trans}, we can see that the estimated result becomes more accurate as the vector number increases. In cases $7\sim 9$, it is noticed that, according to Cannikin Law, the final estimation results are significantly influenced by the worst measurement axis and all the algorithms produce the same behaviors in such cases. Therefore, till now, we can draw the conclusion that the proposed FS3R owns the same accuracy and robustness with SVD and EIG.

\begin{figure*}[hb]
\centering
\includegraphics[width=0.9\textwidth]{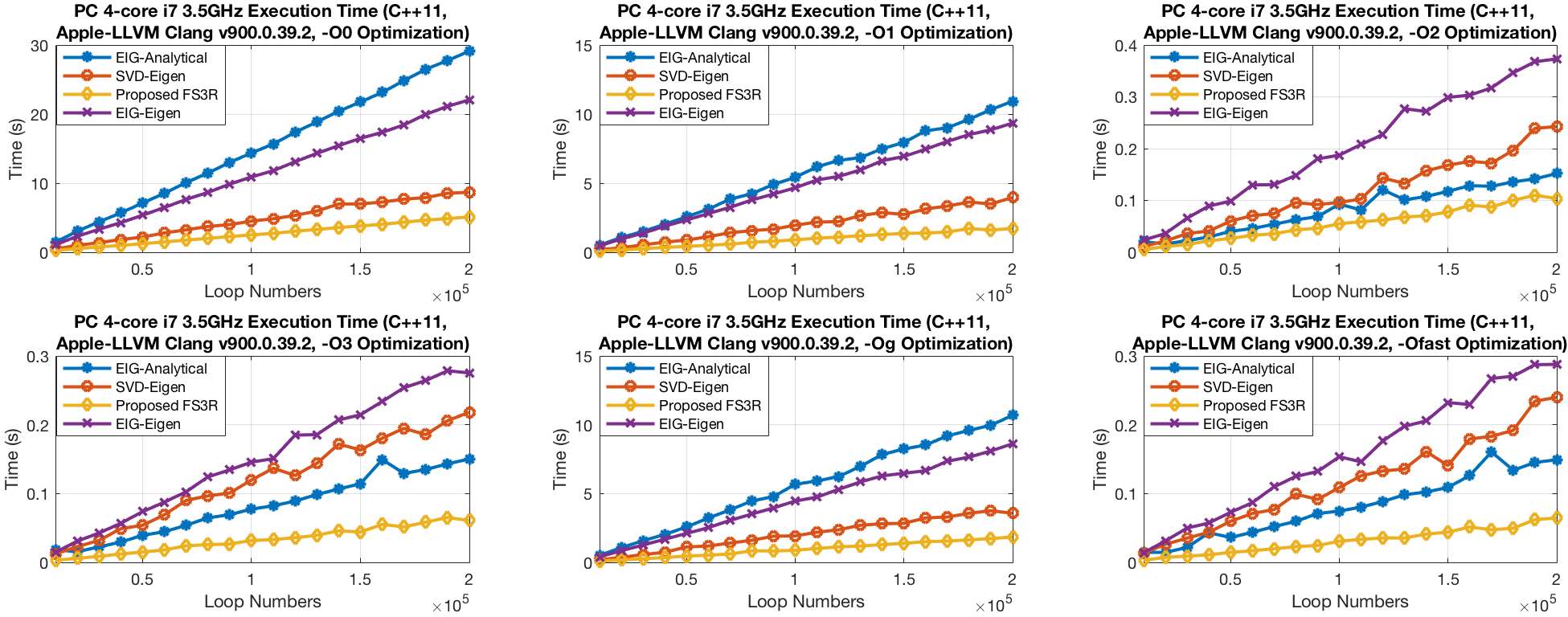}
\caption{Computation time comparisons on the PC.}
\label{fig:time_PC}
\end{figure*}
\begin{figure*}[hb]
\centering
\includegraphics[width=0.9\textwidth]{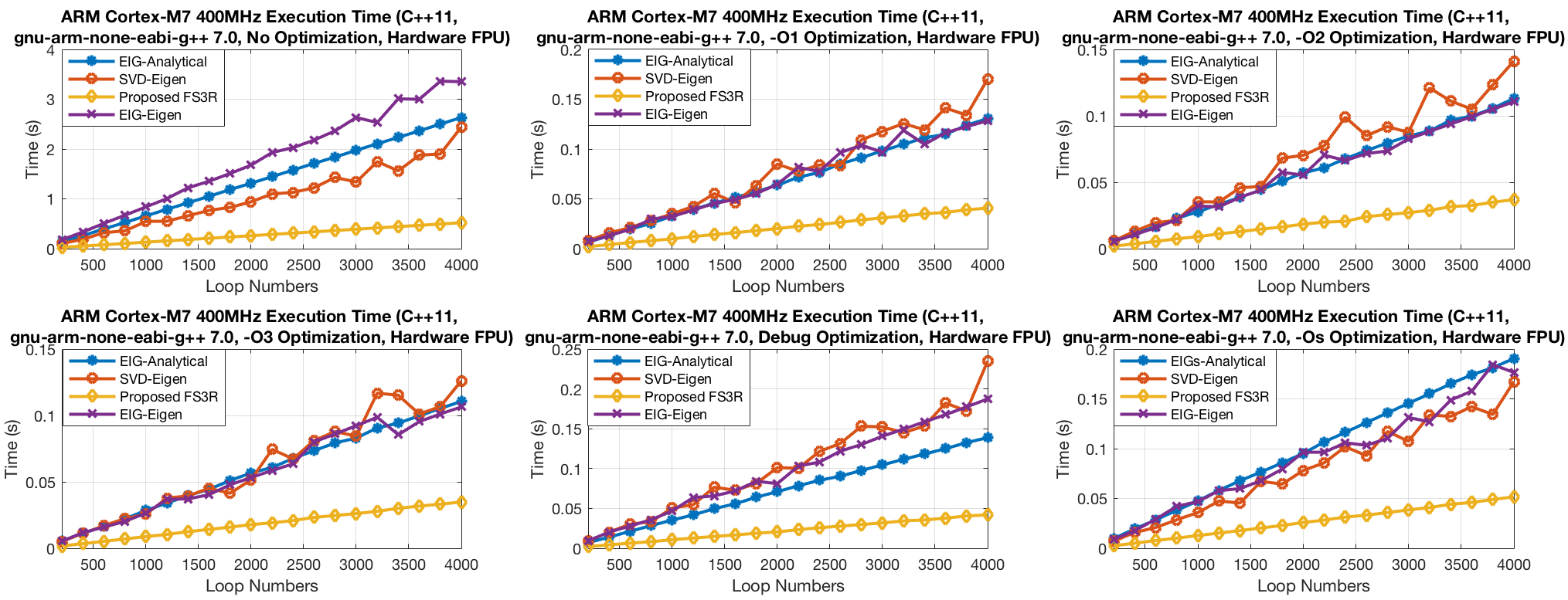}
\caption{Computation time comparisons on the ARM embedded processor.}
\label{fig:time_ARM}
\end{figure*}

\subsection{Computation Time}
The main superiority of the proposed FS3R is that it owns very simple symbolic computation procedure. It is the main reason that it execute very fast in engineering practice. In this sub-section, we rewrite the algorithms 'SVD', 'EIG', 'EIG Analytical' and FS3R using the C++ programming language. They are tested not only on the PC, but on the embedded ARM processor as well. The Eigen matrix computation library is used for matrix manipulations and factorizations. The C++11 programming standard is utilized here ensuring feasible Eigen implementation.\\
\indent For different engineering uses, the developer may choose quite different optimization levels for code generation. Commonly, for high-security productions, the optimization level is relatively low since many optimization options may result in fatal problem in program execution. Hence, we especially evaluate all the algorithms under various optimization levels. The PC is an x64 based laptop with 4-core i7 3.5GHz CPU and the ARM processor is single-chip Cortex-M7 STM32H743VIT6 with clock speed of 400MHz and external FPU for fast double/float number computation. For the PC test, each algorithm is run for 10000 times for averaging execution time. On the ARM processor, as we only have a small RAM area of 1MByte, each algorithm is averaged every 200 cycles. The computation time performances are depicted in Fig. \ref{fig:time_PC} and \ref{fig:time_ARM}.\\
\indent All these algorithms behave with linear time complexity of $O(n)$ but it is obvious that numerical algorithms using Eigen have evident time variance. The main factor is that the stop conditions of such algorithms are usually uncontrollable. While for analytical or symbolic methods, the computation time are quite deterministic. In all the tests, the proposed FS3R shows definite superiority. The time consumption in general takes from $54.43\% $ to $87.12\%$ of existing ones, which is a very large advance that no previous algorithm has reached. The simple procedure of the FS3R saves implementation and compiling time and also decreases the program space. The insurance of the FS3R's accuracy, robustness plus its extremely low computation time makes it a booster in related applications. 

\begin{figure*}[hb]
\centering
\includegraphics[width=0.8\textwidth]{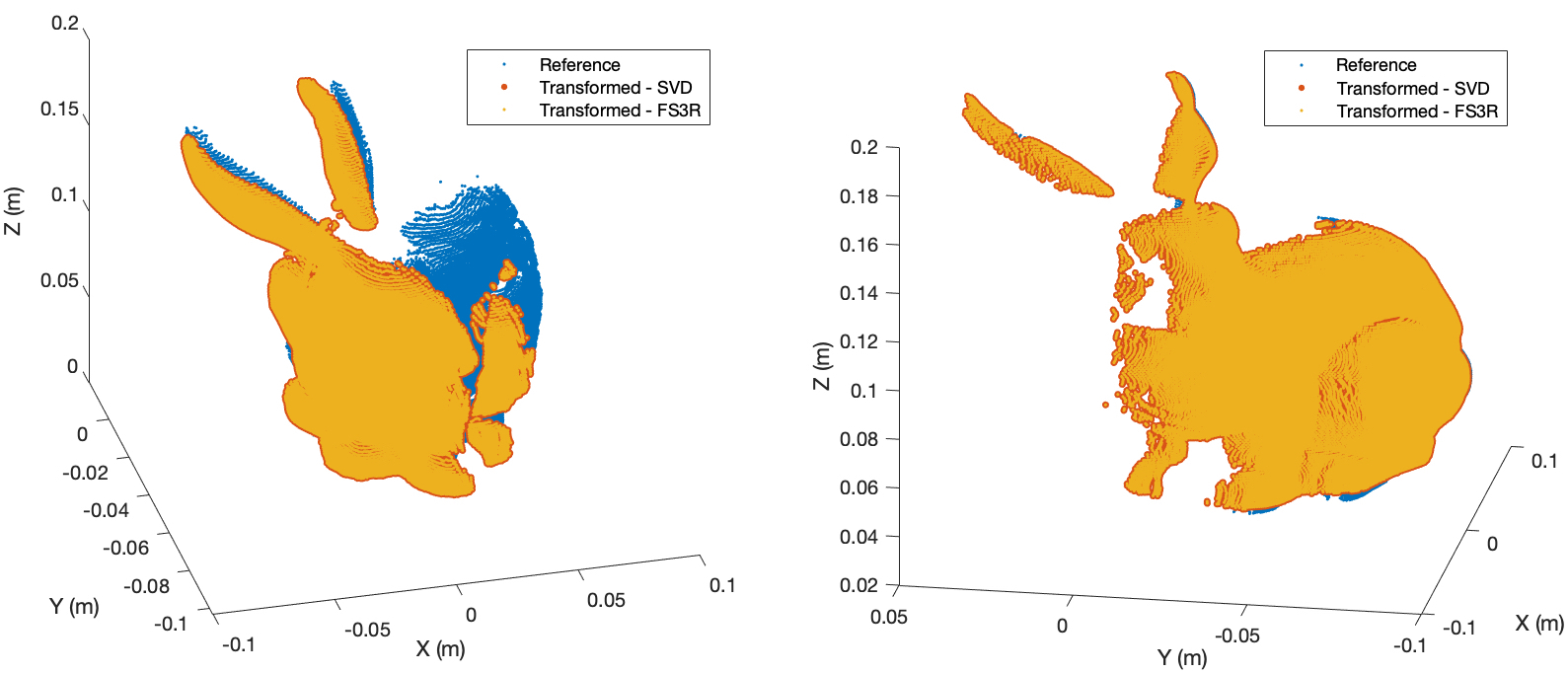}
\caption{The registration results using the 'bunny' dataset from Stanford University \cite{krishnamurthy1996fitting}. Left figure denotes the registration from \texttt{bun270.ply} to \texttt{bun315.ply}; Right figure depicts the registration from \texttt{bun000.ply} to \texttt{bun045.ply}}
\label{fig:bunny}
\end{figure*}
\begin{figure*}[ht]
\centering
\includegraphics[width=0.8\textwidth]{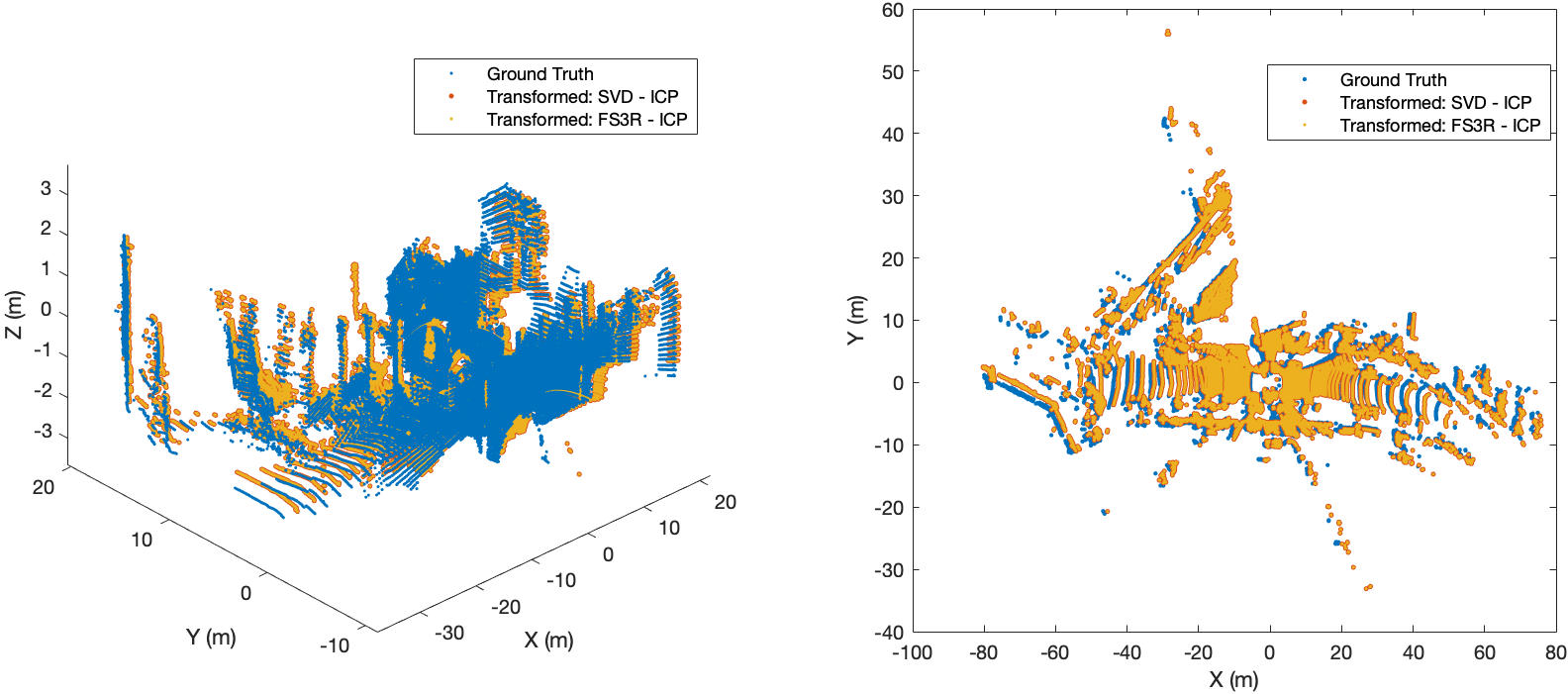}
\caption{The transformations using the KITTI dataset. Left figure: 3D results; Right figure: 2D evaluation performance.}
\label{fig:tform_kitti}
\end{figure*}

\subsection{Evaluation with Open Datasets}
In thi sub-section, the proposed FS3R is introduced for open-dataset evaluation. We pick up two categories of datasets i.e. the 'bunny' dataset from Stanford University \cite{krishnamurthy1996fitting} and the KITTI dataset from Karlsruhe Institute of Technology
and Toyota Technological Institute at Chicago \cite{Geiger2012CVPR}. Both datasets are widely compared in existing literatures \cite{kwok2018dnss,wang2018heading}. The 'bunny' dataset contains multiple-view point-cloud scans of a decorated rabbit model. We use two pairs of correspondences in the 'bunny' dataset to conduct 3D reconstruction using ICP algorithms comprising SVD and proposed FS3R respectively (see Fig. \ref{fig:bunny}). The matching part is implemented using the k-d tree. After 30 iterations, SVD and FS3R converge to ${\mathcal{L}}_{\rm{SVD}}=0.003092129178551$ and ${\mathcal{L}}_{\rm{FS3R}}=0.003092129178549$ respectively. The difference is so tiny that can be ignored in reconstruction of the model. What needs to be point out here is that FS3R only takes $0.56738\ \rm{sec}$ in computation, which is much faster than that $1.13026\ \rm{sec}$ from SVD.
\begin{figure}[H]
\centering
\includegraphics[width=0.48\textwidth]{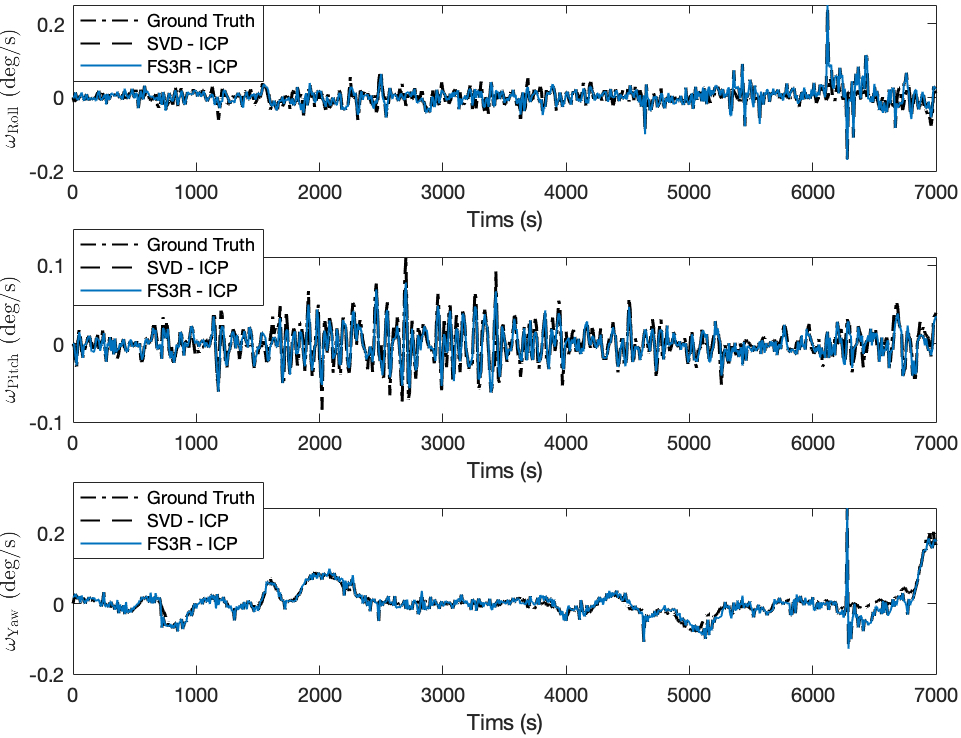}
\caption{Estimated angular rates from KITTI dataset.}
\label{fig:angular_rate_kitti}
\end{figure}
\indent For the KITTI dataset, the source data folder \texttt{2011\_09\_26\_drive\_0014\_sync} is selected. The KITTI dataset has a high-precision ground truth system supported by Velodyne 3D laser scanners, high-end inertial measurement units (IMUs), global positioning system (GPS) receivers and high-resolution color stereo image captures. With point-cloud measurements from the Velodyne laser scanners, the transformation sequence is restored using SVD and proposed FS3R along with the ICP (see Fig. \ref{fig:tform_kitti}). The Euler angles $\varphi ,\vartheta ,\psi$ are converted from rotation matrices from the transformation sequence. With this Euler-angle sequence, the angular rates in three directions roll, pitch and yaw, i.e. $\omega_{\rm{Roll}}, \omega_{\rm{Pitch}}, \omega_{\rm{Yaw}}$ are reconstructed using the tagged timestamps (see Fig. \ref{fig:angular_rate_kitti}).

\begin{figure}[H]
\centering
\includegraphics[width=0.48\textwidth]{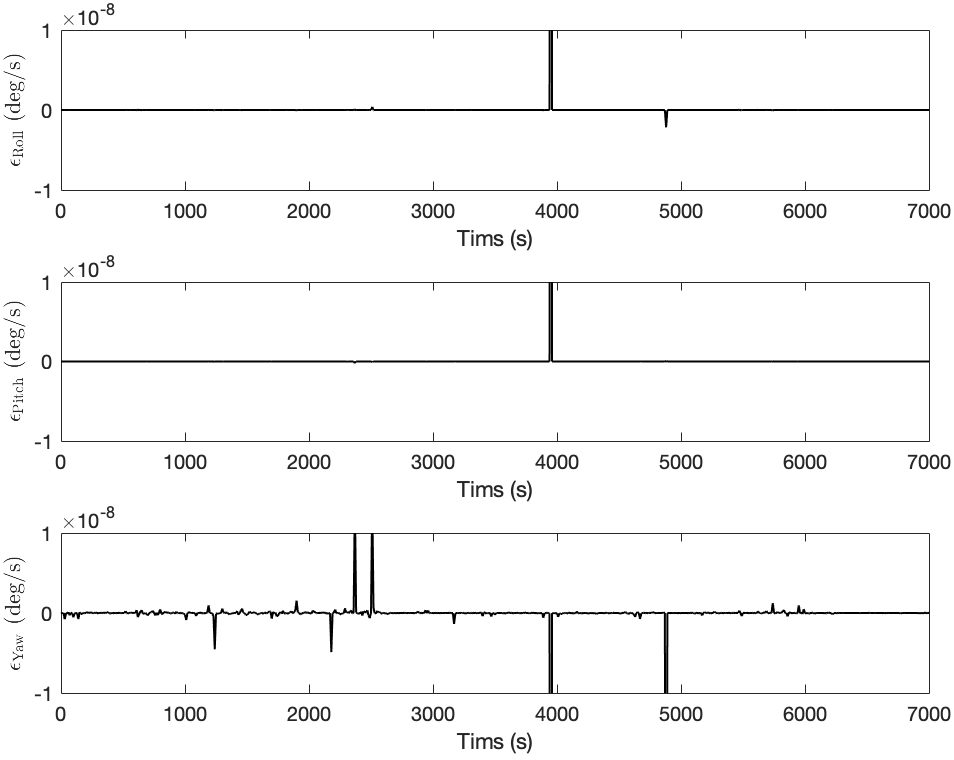}
\caption{Angular-rate error difference between SVD and the proposed FS3R.}
\label{fig:angular_error}
\end{figure}

\begin{figure*}[ht]
\centering
\includegraphics[width=1.0\textwidth]{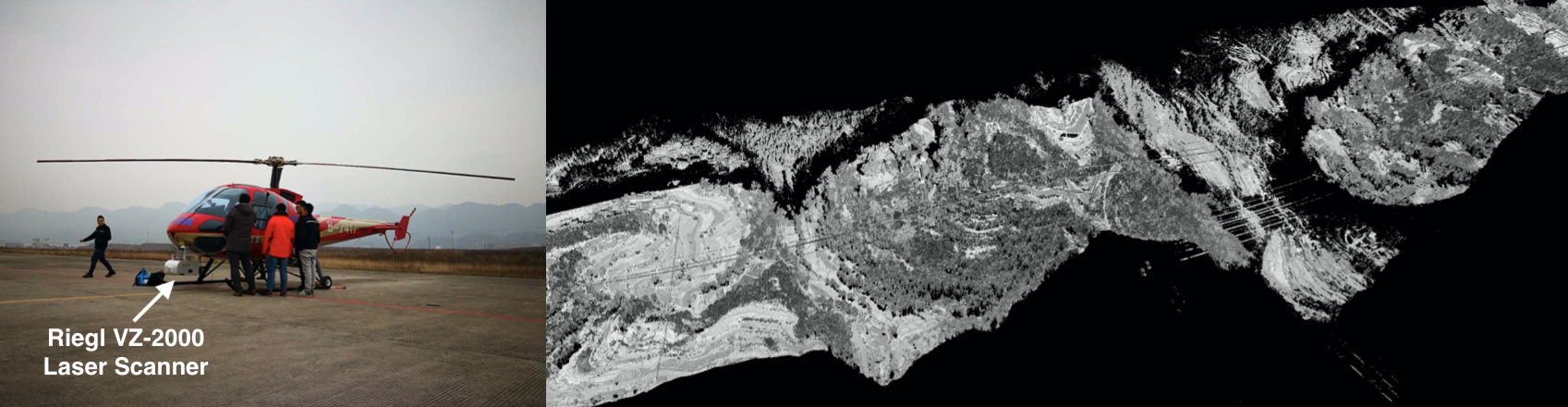}
\caption{Point-cloud capture system on the helicopter (Left) and matching results for power-line inspection using the combination of proposed FS3R and \texttt{libpointmatcher} from ETHZ ASL Lab (Right).}
\label{fig:match}
\end{figure*}

The estimated angular rates from SVD and FS3R are generally the same. Let us define the 'Axis' difference by
\begin{equation}
\begin{gathered}
\epsilon_{\rm{Axis}} = |\omega_{\rm{Axis, SVD}} - \omega_{\rm{Axis, True}}| - |\omega_{\rm{Axis, FS3R}} - \omega_{\rm{Axis, True}}|
\end{gathered}
\end{equation}
where $\rm{Axis} \in \{\rm{Roll, Pitch, Yaw}\}$ and $\omega_{\rm{Axis, True}}$ denotes the true angular rate in the direction of $\rm{Axis}$ obtained from the ground-truth data. We can observe from Fig. \ref{fig:angular_error} that the difference between SVD and proposed FS3R has been enlarged. Here $\epsilon_{\rm{Axis}} > 0$ reflects better performance of FS3R, vice versa. In Fig. \ref{fig:angular_error}, there are more positive peaks than negative ones, which indicates that here FS3R is slightly more accurate than SVD. Note that the scale of such error is in fact enlarged by mass matching process inside the ICP. Here the root mean-squared (RMS) statistics are summarized in Table \ref{tab:kitti}.

\begin{table}[H]
\centering
\caption{RMS results for angular rate estimation from KITTI dataset.}\label{tab:kitti}
\resizebox{0.5\textwidth}{!}{
\begin{tabular}{ccc}
\toprule
{}&{SVD}&{Proposed FS3R}\\
\midrule
{$\epsilon_{\rm{Roll}}$}&{$2.0979 \times 10^{-08}\ \rm{deg/s}$}&{$1.9883 \times 10^{-08}\ \rm{deg/s}$}\\
{$\epsilon_{\rm{Pitch}}$}&{$1.9432 \times 10^{-08}\ \rm{deg/s}$}&{$1.8075 \times 10^{-08}\ \rm{deg/s}$}\\
{$\epsilon_{\rm{Yaw}}$}&{$3.1302 \times 10^{-08}\ \rm{deg/s}$}&{$2.9963 \times 10^{-08}\ \rm{deg/s}$}\\
\bottomrule
\end{tabular}}
\end{table}
The shown errors from FS3R are slightly smaller than that from SVD. FS3R can analytically compute the eigenvalue without iterations and thus will not be influenced by numerical thresholds. Therefore the ICP from FS3R may be more applicable on low-configuration platforms. However, we need to point out that here the error scales in Table \ref{tab:kitti} can almost be ignored as they have nearly reached the level of nano $\rm{deg/s}$, which is better than most of expensive fiber-optic gyroscopes. Such requirements may seldom occur in real engineering practices. Thus here the SVD and proposed FS3R can be regarded to own identical accuracy and robustness.

\subsection{Application Notes}
The FS3R, since its invention, has been applied to some time-consuming tasks e.g. point-cloud registration and video stitching. In a recent test where a huge point cloud containing 7889456 points are captured using the Riegl VZ-2000 3D laser scanner on a real-world helicopter (see Fig. \ref{fig:match}) for power-line inspection. The original registration method is motivated by the \texttt{libpointmatcher} from ETHZ ASL Lab \cite{Pomerleau12comp} in which the ICP is completed using the SVD. By replacing SVD with the proposed FS3R, the matching time has been decreased from 1 hour to 0.65 hour. The general results of the point matching are also shown in Fig. \ref{fig:match}. The SVD and FS3R in real engineering use have been verified to own the same accuracy and robustness. The huge amount of point correspondences ensure precision estimates of rigid-body transformation. This shows the potential applicability of FS3R in industrial processing. We also have made all the codes in C++ and MATLAB open-source and the audience can verify its effectiveness (see the footnote of the first page).

\section{Conclusion}
Our recent algorithm FLAE is revisited, which is later related to the 3D registration problem. Some proofs are presented to show the equivalence. The previous solution to quartic equation is then simplified getting rid of complex numbers for easier implementation. Numerical robustness of the proposed method is also investigated showing its immunity to degenerated matrices. The proposed algorithm is systematically evaluated with other representatives. The results indicate that it maintains the accuracy and robustness but consumes much less computation time. Real applications including large-point-cloud registration have shown its superiority in engineering processing. However, it is still noticed that the current method highly relies on the floating-point operations. Unless we have reached the limit of EIG numerical algorithm, we still have an expectation to develop the next-generation algorithm in which the floating-number is no longer need, which, would be of great benefit for parallel computing platforms like FPGA and GPU for accelerated performance.



%

%

\section*{Acknowledgment}
This research was supported by Shenzhen Science, Technology and Innovation Comission (SZSTI) JCYJ20160401100022706, awarded to Prof. Ming Liu., in part by National Natural Science Foundation of China under the grant of No. 41604025, in part by General Research Fund of Research Grants Council Hong Kong,11210017, and also in part by Early Career Scheme Project of Research Grants Council Hong Kong, 21202816 The authors would also like to thank Dr. F. Landis Markley from NASA Goddard Space Flight Center and Prof. Yuanxin Wu from Shanghai Jiao Tong University for their constructive comments to this paper. Dr. Yaguang Yang from Nuclear Research Centre, U.S. Navy provided his codes of \cite{Yang2013}. The experimental devices e.g. laser scanner and helicopter are supported by Chongqing Fengmai Innovation Inc., Chongqing, China. We genuinely thank them for their help. The source code of this paper has been uploaded to https://github.com/zarathustr/FS3R (\textbf{C++}), https://github.com/zarathustr/FS3R-Matlab (\textbf{MATLAB}) and https://github.com/zarathustr/FS3R-CrossWorks (\textbf{Embedded}).

\ifCLASSOPTIONcaptionsoff
  \newpage
\fi



\ \\
\ \\
\ \\
\ \\
\ \\
\ \\
\ \\
\ \\
\ \\
\ \\
\ \\
\ \\
\ \\
\ \\
\ \\
\ \\
\ \\
\ \\

\begin{IEEEbiography}[{\includegraphics[width=1in,height=1.25in,clip,keepaspectratio]{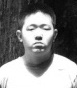}}]{Jin Wu} was born in May, 1994 in Zhenjiang, China. He received the B.S. Degree from University of Electronic Science and Technology of China, Chengdu, China. He has been a research assistant in Department of Electronic and Computer Engineering, Hong Kong University of Science and Technology since 2018. His research interests include robot navigation, multi-sensor fusion, automatic control and mechatronics. He is a co-author of over 30 technical papers in representative journals and conference proceedings of \textsc{IEEE, AIAA, IET} and etc. Mr. Jin Wu received the outstanding reviewer award for \textsc{Asian Journal of Control}. One of his papers published in \textsc{IEEE Transactions on Automation Science and Engineering} was selected as the ESI Highly Cited Paper by ISI Web of Science during 2017 to 2018. He has been in the UAV industry from 2012 and has launched two companies ever since. He is a member of IEEE.
\end{IEEEbiography}

\begin{IEEEbiography}[{\includegraphics[width=1in,height=1.25in,clip,keepaspectratio]{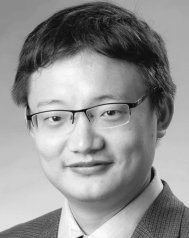}}]{Ming Liu}
received the B.A. degree in
automation from Tongji University, Shanghai, China,
in 2005, and the Ph.D. degree from the Department of Mechanical and Process Engineering, ETH
Zürich, Zürich, Switzerland, in 2013, supervised by
Prof. Roland Siegwart. During his master’s study
with Tongji University, he stayed one year with the
Erlangen-Nünberg University and Fraunhofer Institute IISB, Erlangen, Germany, as a Master Visiting
Scholar.\\
\indent He is currently with the Electronic and Computer
Engineering, Computer Science and Engineering Department, Robotics Institute, The Hong Kong University of Science and Technology, Hong Kong, as an Assistant Professor.
He is also a founding member of Shanghai Swing Automation Ltd., Co.
He is coordinating and involved in NSF Projects and National 863-Hi-TechPlan Projects in China. His research interests include dynamic environment
modeling, deep-learning for robotics, 3-D mapping, machine learning, and
visual control.\\
\indent Dr. Liu was a recipient of the European Micro Aerial Vehicle Competition
(EMAV’09) (second place) and two awards from International Aerial Robot
Competition (IARC’14) as a Team Member, the Best Student Paper Award
as first author for MFI 2012 (IEEE International Conference on Multisensor
Fusion and Information Integration), the Best Paper Award in Information
for IEEE International Conference on Information and Automation (ICIA
2013) as first author, the Best Paper Award Finalists as co-author, the Best
RoboCup Paper Award for IROS 2013 (IEEE/RSJ International Conference
on Intelligent Robots and Systems), the Best Conference Paper Award for
IEEE-CYBER 2015, the Best Student Paper Finalist for RCAR 2015 (IEEE
International conference on Real-time Computing and Robotics), the Best
Student Paper Finalist for ROBIO 2015, the Best Student Paper Award for
IEEE-ICAR 2017, the Best Paper in Automation Award for IEEE-ICIA 2017,
twice the innoviation contest Chunhui Cup Winning Award in 2012 and 2013,
and the Wu Wenjun AI Award in 2016. He was the Program Chair of IEEERCAR 2016 and the Program Chair of International Robotics Conference in
Foshan 2017. He was the Conference Chair of ICVS 2017. He has published many popular papers in top robotics journals including \textsc{IEEE Transactions on Robotics}, \textsc{International Journal of Robotics Research} and \textsc{IEEE Transactions on Automation Science and Engineering}. Dr. Liu is currently an Associate Editor for \textsc{IEEE Robotics and Automation Letters}. He is a Senior Member of IEEE.
\end{IEEEbiography}

\begin{IEEEbiography}[{\includegraphics[width=1in,height=1.25in,clip,keepaspectratio]{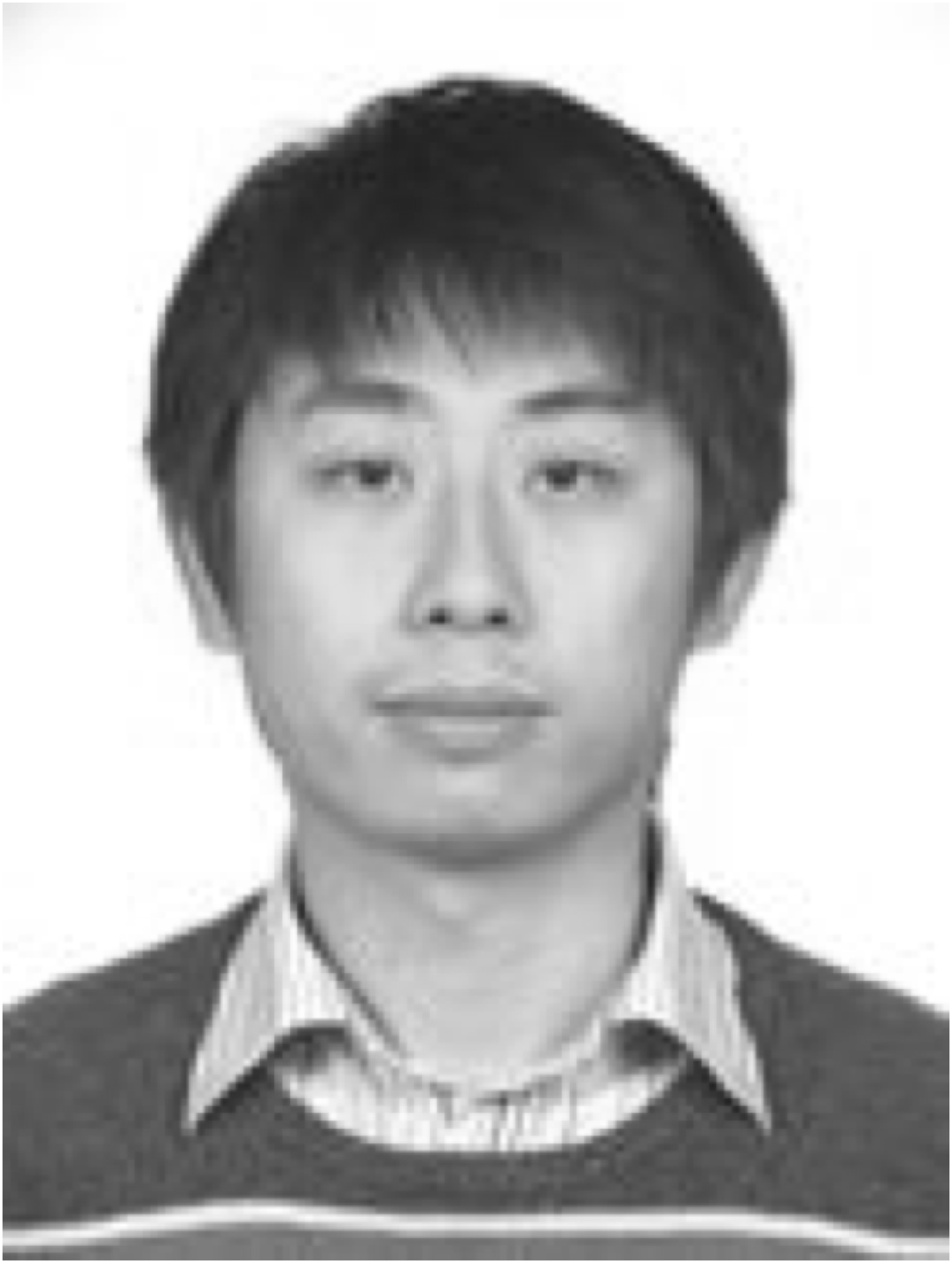}}]{Zebo Zhou} was born in November, 1982 in Yongchuan, Chongqing, China. He received the B.Sc. and M.Sc. degrees in School of Geodesy and Geomatics from Wuhan University, Wuhan, China, in 2004 and 2006, respectively, and the Ph.D. degree from the College of Surveying and Geoinformatics, Tongji University, Shanghai, China in 2009. He was a visiting fellow with the Surveying \& Geospatial Engineering Group, within the School of Civil \& Environmental Engineering, University of New South Wales, Australia in 2009 and 2015. He is currently an associate professor with the School of Aeronautics and Astronautics, University of Electronic Science and Technology of China, Chengdu, China. His research interests include GNSS navigation and positioning, GNSS/INS integrated navigation, multi-sensor fusion.\\
\indent Prof. Zhou has been in charge of projects of National Natural Science Foundation of China and has taken part in the National 863 High-tech Founding of China. He served as a Guest Editor of several special issues published on \textsc{International Journal of Distributed Sensor Networks} and \textsc{Asian Journal of Control}. He has been presenting related works on the annual conference of the institute of navigation (ION), the annual  Chinese satellite navigation conference (CSNC) for several times and has received the best paper awards in these conferences.
\end{IEEEbiography}

\begin{IEEEbiography}[{\includegraphics[width=1in,height=1.25in,clip,keepaspectratio]{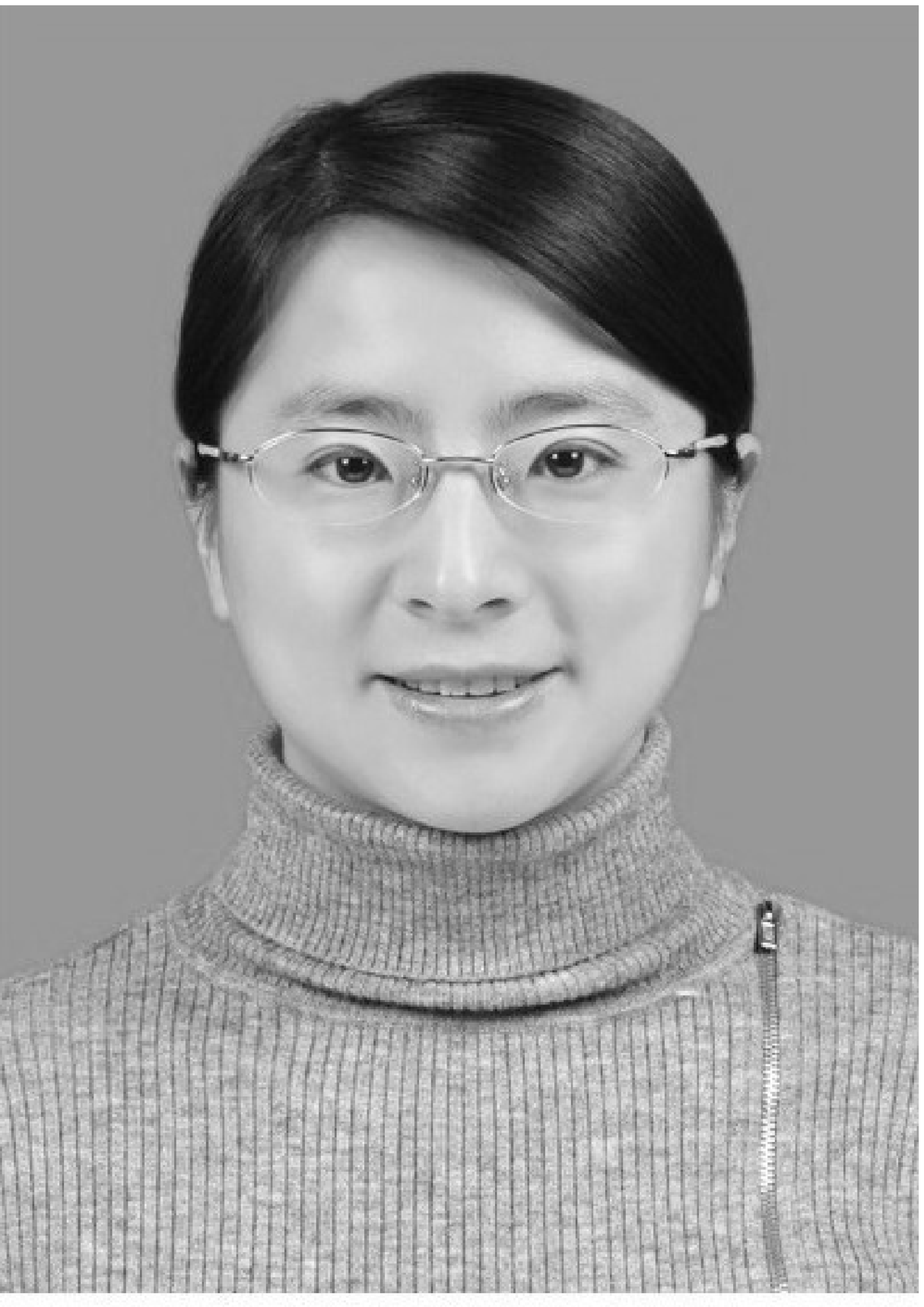}}]{Rui Li}
received the Ph.D. degree in Control Science and Engineering from Harbin Institute of Technology, China, in 2008. She joined University of Electronic Science and Technology of China (UESTC) in 2008 where she is currently an associate professor in School of Automation, UESTC. Previously, she worked as a Visiting Research Associate with Department of Applied Mathematics, the Hong Kong Polytechnic University and as Visiting Research Associate with the Department of Mathematics and Statistics, Curtin University of Technology. From September 2011 to September 2012, she was a visiting scholar with Department of Electrical Engineering, University of California at Riverside. Her research interests include optimization theory and optimal control, nonlinear control, multi-agent systems and aircraft control. She is a member of IEEE.
\end{IEEEbiography}

\end{document}